\documentclass{llncs}

\usepackage{etex}
\usepackage{times}
\usepackage{graphicx}
\usepackage{float}
\usepackage{latexsym}
\usepackage{url}
\usepackage[ruled,vlined]{algorithm2e}
\usepackage{pgf}
\usepackage[utf8]{inputenc}

\usepackage{xcolor}
\usepackage{colortbl}
\usepackage{soul}
\definecolor{darkblue}{rgb}{0,0,0.75}

\usepackage{amsmath}
\usepackage{amssymb}
\usepackage{latexsym,xspace}
\usepackage{graphics,graphicx,float}
\usepackage{times,psfrag}
\usepackage{rotate}
\usepackage{epsfig}
\usepackage{subfigure}

\usepackage{tikz}
\usetikzlibrary{arrows,shapes,snakes,automata,backgrounds,petri,shadows}

\def\viewResul#1{\cellcolor[gray]{0.85} #1}

\title{Improving \muc extraction thanks to \\ local search}
\author{
  \'Eric Grégoire
  \and
  Jean-Marie Lagniez  
  \and
  Bertrand Mazure
}

\institute
{CRIL \\
  Universit\'e d'Artois \&
  CNRS, UMR 8188\\
  rue Jean Souvraz SP18 \\
  F-62307 Lens~  France\\
  \email{\{gregoire,lagniez,mazure\}@cril.fr}  
}

\newcommand{\sat}{\textsc{sat}\xspace}

\newcommand{\dc}{\textsc{dc}\xspace}
\newcommand{\mr}{\textsc{mr}\xspace}
\newcommand{\wcore}{\textsc{wcore}\xspace}
\newcommand{\mac}{\textsc{mac}\xspace}
\newcommand{\muc}{\textsc{muc}\xspace}
\newcommand{\mus}{\textsc{mus}\xspace}
\newcommand{\cn}{\textsc{cn}\xspace}
\newcommand{\csp}{\textsc{csp}\xspace}
\newcommand{\CSP}{\textsc{csp}\xspace}

\newcommand{\affect}{\leftarrow}

\newcommand{\SLS}{\textsc{sls}\xspace}

\newcommand{\dcwcore}{\textsc{dc}(\wcore)}
\newcommand{\rclcsp}{\textsc{rclCsp}\xspace}
\newcommand{\LSTC}{\textsc{lstc}}
\newcommand{\rmr}{\textsc{r}ec-\mr}

\newcommand{\Ass}{\mathcal{A}\xspace}
\newcommand{\A}{\mathcal{A}\xspace}
\newcommand{\Ccut}{\mathcal{C}_{\textsc{cut}}\xspace}
\newcommand{\ALS}{\mathcal{A}_{\textsc{tr}}\xspace}
\newcommand{\dom}{dom}
\newcommand{\scp}{scp}

\newcommand{\CN}{constraint network\xspace}

\newcommand{\CNLR}[2]{\langle #1, #2 \rangle\xspace}
\newcommand{\PXC}{\mathcal{P}=\langle\X,\C\rangle}
\newcommand{\XC}{\langle\X,\C\rangle}
   
\newcommand{\C}{\mathcal{C}}
\renewcommand{\P}{\mathcal{P}}
\newcommand{\X}{\mathcal{X}}

\begin{document}

\maketitle

\begin{abstract}
  Extracting \muc{s} (Minimal Unsatisfiable Cores) from an
  unsatisfiable constraint network is a useful process when causes of
  unsatisfiability must be understood so that the network can be
  re-engineered and relaxed to become satisfiable.  Despite bad worst-case
  computational complexity results, various \muc-finding approaches that appear tractable
  for many real-life instances have been proposed.  Many of them are
  based on the successive identification of so-called transition
  constraints.  In this respect, we show how local search can be used to
  possibly extract additional transition constraints at each main iteration step. 
  The approach is shown to outperform a  technique based on a form
  of model rotation imported from the \sat-related technology and that
  also exhibits  additional transition
  constraints.
   Our extensive computational experimentations show
  that this enhancement also boosts the performance of state-of-the-art
  \dc{(\wcore{})}-like \muc extractors.
  \end{abstract}

% ------------------------------------------------------------------------- 
\section{Introduction}

In this paper, the focus is on unsatisfiable constraint networks. More
precisely, a new approach for  extracting minimal cores (or, \muc{s} for
\textit{Minimally Unsatisfiable Cores}) of constraint networks is
proposed.  A \muc is a minimal (w.r.t. $\subseteq$) set of constraints
that cannot be satisfied all together.  When causes of 
unsatisfiability must be understood and the network must be
re-engineered and relaxed to become satisfiable, extracting \muc{s} can be
a cornerstone issue since a \muc provides one explanation for
unsatisfiability in terms of a minimal set of incompatible
constraints.  Despite bad worst-case computational complexity results, various
approaches for  extracting one \muc have been proposed that appear
tractable for many  instances
\cite{siquiera88,bakker-etal93,choco,jussien00a,junker01,junker04,Hemery2006,GMP+-06-1,GMP-08-4}.

A \muc of a network can also be defined as an unsatisfiable
sub-network formed of transition constraints, which are constraints
that would allow this sub-network to become satisfiable if any of them
were removed.  Powerful approaches to \muc extraction are founded on
transition constraints, both in the \csp \cite{Hemery2006,GMP-08-4}
and the \sat
\cite{Lynce-Joao:04,GMP-06-3,GMP-07-2,VanMaaren-Wieringa:08,Lynce-Joao:04,Ryvchin-Strichman:11,Belov-Joao:12}
domains. In this last area, a recent approach
\cite{Lynce-Joao:11,Belov2011} focuses on the following intuition. An
assignment of values to the variables that satisfies all constraints
except one is called a transition assignment and the unsatisfied
constraint is a transition constraint: additional transition
constraints might be discovered by so-called model rotation, i.e., by
examining other assignments differing from the transition assignment
on the value of one variable, only.

In the paper, an approach both extending and enhancing this latter
technique is proposed in the constraint network framework. The main
idea is to use local search for exploring the neighborhood of
transition assignments in an attempt to find out other transition
constraints.  The technique is put in work in a so-called dichotomy
destructive strategy \textit{\`{a} la} \dc{(\wcore{})}
\cite{Hemery2006} to extract one \muc.  Extensive computational
experimentations show that this approach outperforms both the model
rotation technique from \cite{Belov2011} and the performance of
state-of-the-art \muc extractors.

The paper is organized as follows. In the next section, basic
concepts, definitions and notations are provided. Section 3 focuses on
existing techniques for \muc extraction, including
\dc{(\wcore{})}-like ones, and then on model rotation.  In section 4,
a local search procedure for exhibiting additional transition constraints is presented and
motivated, whereas section 5 describes the full algorithm for \muc
extraction.  Section 6 describes our experimental investigations and
results before some promising paths for further research are
presented in the conclusion.

%%%%%%%%%%%%%%%%%%%%%%%%%%%%%%%%%%%%%%%%%%%%%%%%%%%%%%%%%%%%%%%%%%%%%%%%%%%% 
\section{Definitions and Notations}

Constraint networks are defined as follows.

\begin{definition}[Constraint network]
  A constraint network (\cn) is a pair $\CNLR{ \X }{ \C }$, where
  \begin{enumerate}
  \item $\X$ is a finite set of variables s.t. each variable $x\in\X$
    has an associated finite instantiation domain, denoted $\dom(x)$;
    % which thus defines  the set of instantiation  values for  $x$;
  \item $\C$ is a finite set of constraints s.t. each constraint
    $c\in\C$ 
    involves a subset of variables of $\X$, called scope and denoted $\scp(c)$. $c$ translates 
    an associated relation %$rel(c)$,
    that  contains all the values  for  the variables of $\scp(c)$ that satisfy $c$.
  \end{enumerate}
\end{definition}

A \CN where the scope of all constraints is binary can be represented
by a non-oriented graph where each variable is a node and each
constraint is an edge. %, labelled with its corresponding relation. 

\begin{figure}[t]
  \centering
  \null\hfill
  \subfigure[\label{fig:csp}  \CN of Example 1.]{
    \scalebox{1}
    {
      \begin{tikzpicture}
        \tikzstyle{var}=[circle,draw]
        \tikzstyle{con}=[midway,sloped,above]
        
        % les connecteurs      
        \node[var] (i) at (0,0.7) {$i$};
        \node[var] (l) at (6,3.3) {$l$};
        \node[var] (m) at (0,3.3) {$m$};
        \node[var] (j) at (6,0.7) {$j$};
        \node[var] (k) at (3,1.9) {$k$};

        \draw (i) -- (m) node[con] {$c_1:m>i$};
        \draw (m) -- (l) node[con] {$c_2:m=l+2$};
        \draw (j) -- (l) node[con,below] {$c_3:j\leqslant l$};
        \draw (i) -- (j) node[con,below] {$c_4:i<j$};
        \draw (i) -- (k) node [con] {$c_5:k<i$};
        \draw (k) -- (j) node [con] {$c_6:j<k$};
        \draw (k) -- (l) node [con] {$c_7:k\ne l~$};
      \end{tikzpicture}
    }
  }
  \hfill
  \subfigure[\label{fig:muc} one \muc in Example 1.]{
    \scalebox{1}
    {
      \begin{tikzpicture}
        \tikzstyle{var}=[circle,draw]
        \tikzstyle{con}=[midway,sloped,above]
        
        % les connecteurs      
        \node[var] (i) at (0,0) {$i$};
        \node[var] (j) at (3,0) {$j$};
        \node[var] (k) at (1.5,2) {$k$};

        \draw (i) -- (j) node[con,below] {$c_4:i<j$};
        \draw (i) -- (k) node [con] {$c_5:k<i$};
        \draw (k) -- (j) node [con] {$c_6:j<k$};
      \end{tikzpicture}
    }
  }
  \hfill\null
  \caption{Graphical representation of  Example 1.}
\end{figure}
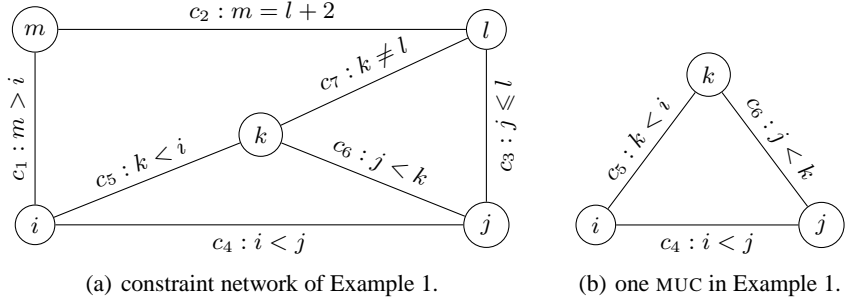

\begin{example}\label{ex:csp}
  Let $\X=\{i,j,k,l,m\}$ where each variable has the same domain $\{0,
  1, 2, 3, $ $ 4\}$ and let $\C=\{c_1:m>i,c_2:m=l+2,c_3:j\geqslant
  l,c_4:i<j,c_5:k<i,c_6:j<k,c_7:c\ne l\}$ be a set of $7$
  constraints. The \CN $\P = \CNLR{ \X }{ \C }$ can be represented by
  the graph of Fig. \ref{fig:csp}.
\end{example}

\begin{definition}[Assignment and solution]
  An assignment $\A$ of a \CN $\PXC$ is an assignment of values to all
  variables of $\X$.  A solution to $\P$ is any assignment that
  satisfies all constraints of $\C$.
\end{definition}

A form of  Constraint Satisfaction Problem (\textsc{csp}) consists in
checking whether a constraint network $\P$ admits at least one
solution.  This decision problem is an {\sl NP-complete} problem. If $\P$
admits at least one solution then $\P$ is called satisfiable else $\P$
is called unsatisfiable. The constraint network of Example
\ref{ex:csp} is unsatisfiable. When a constraint network is
unsatisfiable, it admits at least one \textit{Minimal
  (w.r.t. $\subseteq$) Unsatisfiable Core} (in short, \muc).

\begin{definition}[Core and \muc]
  Let $\PXC$ be a constraint network. \\
  $\P'=\langle\X',\C'\rangle$ is 
  an unsatisfiable core, in short a core, of $\P$ iff
  \begin{itemize}
  \item $P'$ is  an unsatisfiable constraint network, and
  \item $\X' \subseteq \X$ and $\C'\subseteq \C$.
  \end{itemize} 

  \noindent
  $P'$ is a Minimal Unsatisfiable Core (\muc) of $P$ iff
  \begin{itemize}
  \item $P'$ is a core of $P$, and
  \item there does not exist any proper core of $P'$: $\forall c\in C',
    \langle \X',C'\setminus\{c\}\rangle$ is satisfiable.
  \end{itemize} 
\end{definition} 

\noindent
\textit{Example 1. (cont'd)}
$\P$ is unsatisfiable and admits only one \muc, 
illustrated in Fig.  \ref{fig:muc}.

Whenever $\P$ is unsatisfiable, $\P$ exhibits at least one \muc. In
the worst case, there can be a number of different \muc{s} that is
exponential in the number $m$ of constraints of $\P$ (actually it is
in $\mathcal{O}(C_{m}^{m/2}))$.  Note that different \muc{s} of a same
network can share constraints.  Accordingly, all \muc{s} need not be extracted in a step-by-step
relaxation process to make the network become satisfiable. Especially, such  an iterative process where
at each step  one \muc is 
extracted and relaxed so that it becomes  satisfiable,  at most $\mathcal{O}(m)$ \muc{s} 
need to be extracted.

\begin{example}
  \label{ex:mucs}
  Fig. \ref{fig:mucs} depicts an unsatisfiable \CN with three \muc{s},
  namely $\{c_1, c_2, c_3\}$, $\{c_3, c_4, c_5\}$ and $\{c_1, c_2,
  c_4, c_5\}$. In this example, each variable is given the same domain
  $\{1,2\}$.
  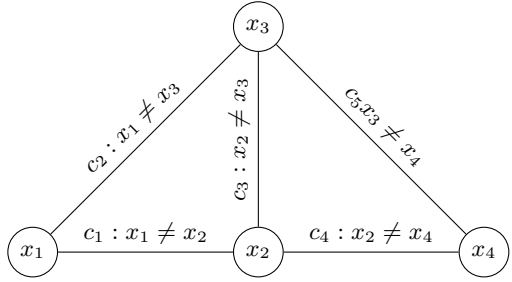
\begin{figure}[t]
    \centering
    \scalebox{1}
    {
      \begin{tikzpicture}

        \tikzstyle{var}=[circle,draw]
        \tikzstyle{con}=[midway,sloped,above]
        
        % les connecteurs      
        \node[var] (x1) at (0,0) {$x_1$};
        \node[var] (x2) at (3,0) {$x_2$};
        \node[var] (x4) at (6,0) {$x_4$};
        \node[var] (x3) at (3,3) {$x_3$};

        \draw (x1) -- (x2) node[con] {$c_1:x_1\ne x_2$};
        \draw (x1) -- (x3) node[con] {$c_2:x_1\ne x_3$};
        \draw (x2) -- (x3) node[con] {$c_3:x_2\ne x_3$};
        \draw (x2) -- (x4) node[con] {$c_4:x_2\ne x_4$};
        \draw (x3) -- (x4) node[con] {$c_5x_3\ne x_4$};
      \end{tikzpicture}
    }
    \caption{\label{fig:mucs} An unsatisfiable constraint network
      that contains 3 \muc{s}.}
  \end{figure}
\end{example}

Extracting   one  \muc from an unsatisfiable constraint network is a heavy computational task in the worst
case. Indeed, checking whether a \CN is a \muc is {\sl DP-complete} \cite{Papadimitriou-Wolfe:88}.
Despite the aforementioned bad worst-case computational complexity property,  various families of approaches
that run in acceptable time  for many 
instances have been proposed.  We describe representatives of some of the main ones  in the next section.

\section{\muc Extraction}

Most  recent approaches for  extracting one  \muc from an unsatisfiable  \CN $\P$ start by
computing one core (which does not need to be  minimal) of $\P$. 
This step can be optional  because an unsatisfiable \CN is already a
core. During this step some information can however be  collected that will  help guiding 
the further minimization step. 
In this paper, we focus on this  minimization step and make use  of the 
\wcore core extractor introduced in \cite{Hemery2006} as a preprocessing step, that we briefly describe
hereafter.

\subsection{ \wcore as a pre-processing step}

When the unsatisfiability of a \CN $\P$ is proved thanks to a
filtering search algorithm, \wcore \cite{Hemery2006} delivers a core
of $\P$ that is formed of all the constraints that have been involved
in the proof of unsatisfiability, namely all the constraints that have
been used during the search to remove by propagation at least one
value from the domain of any variable.  Such constraints are called
\textit{active}.  Therefore, when $\P$ is shown unsatisfiable, active
constraints form a core since the other constraints do not actually
take part to this proof of inconsistency. 
The approach from \cite{Hemery2006}  iterates this process until
no smaller set of active constraints is found.
Consequently, at the end of this first
step, constraints that are not active can be removed safely while
keeping a remaining \CN that is unsatisfiable.

Clearly, the resulting core can depend on the order according to which
the partial assignments are investigated, which is guided by some
branching heuristic.  \wcore takes advantage of the powerful
\emph{dom/wdeg} heuristic \cite{Boussemart2004} (see also variants in
e.g. \cite{GMP-08-4}), which consists in attaching to each constraint
a counter initialized to $1$ and that is incremented each time the
corresponding constraint is involved in a conflict. In this respect,
\emph{dom/wdeg} selects the variable with the smallest ratio between
the current domain size and a weighted degree, which is defined as the
sum of the counters of the constraints in which the variable is
involved.  This heuristic allows to focus on constraints that appear
difficult to satisfy. The goal is not only to attempt to ease the
search for inconsistency but also to record some indication that these
constraints are probably prone to belong to a \muc.  Accordingly, it
is proposed in \cite{Hemery2006} to weigh the constraints via the
\emph{dom/deg} heuristic and use the \wcore approach as a
preprocessing step for \muc extraction to attempt to reduce the size
of the core. Likewise, our approach reuses this weighing information
in the subsequent steps of the algorithm to compute one \muc.

\subsection{Minimization step}

Once a core has been extracted from a \CN, it must be minimized so
that if forms one \muc. To this end, it might be necessary to check
whether a constraint belongs or not to the set of \muc{s} included
within a core, which is a task in $\Sigma^p_2$
\cite{Eiter-Gottlob:92}.  In practice, this step is often based on the
identification of forms of \emph{transition constraints}.

\begin{definition}[Transition constraint]
  Let $\P =\CNLR{\X}{\C}$ an unsatisfiable constraint network.  $c \in
  \C$ is a transition constraint of $\P$ iff there exists an
  assignment $\Ass$ of $\P$ such that $\Ass$ is a model of
  $\CNLR{\X}{\C\setminus\{c\}}$.
\end{definition}

\noindent
\textit{Example 1 (cont'd)} Consider again Fig. \ref{fig:csp}.
$c_{4}$ is a transition constraint.  Indeed $\P$ is unsatisfiable and
$\Ass = \{i=2,j=0,k=1,l=2,m=4\}$ is a solution of
$\CNLR{\X}{\C\setminus\{c_{4}\}}$.

The following property is straightforward and directly follows from the definition of transition constraints.
\begin{property}
  If $c$ is a transition constraint of a core $\P$ then $c$ belongs to
  any \muc of $\P$.
\end{property}

Clearly, all \muc{s} of a core do not necessarily share a non-empty
intersection and a \CN might thus have no transition constraints.
Actually, the process of finding out transition constraints is
performed with respect to some subparts of the network adopting
e.g. either so-called destructive or constructive approaches
\cite{siquiera88,bakker-etal93,jussien00a,junker01,Hemery2006,GMP-08-4}. For
example, the \emph{constructive} approaches (as in \cite{siquiera88})
successively insert constraints taken from the core into a set of
constraints until this latter set becomes unsatisfiable.  At the
opposite, \emph{destructive} approaches \cite{bakker-etal93}
successively remove constraints from the initial core until the
current network becomes satisfiable. Constraints are ordered and each
time a transition constraint is discovered, it is placed at the
beginning of the core according to this order. All constraints are
tested according to the inverse order.  It is also possible to use a
\emph{dichotomy strategy} in order to find out transition constraints
\cite{Hemery2006}. Variants and combinations of these techniques can
be traced back to e.g. \emph{QuickXplain} \cite{junker04,Joao2013} and
the \emph{combined approach} \cite{GMP-08-4}.

Clearly, the order according to which the constraints are tested is
critical for the efficiency of each approach. This order can be set
according to the weighs of constraints computed during the \wcore
step.  In the rest of the paper, we focus on a \emph{dichotomy
  destructive} approach, which will be presented in more detail later
on.  Before that, let us briefly present a method that has been
recently proposed in the \sat research community to find more than one
transition constraint at each main iteration of a \muc extraction
algorithm (in the \sat domain, a \muc is called a \mus for
\textit{Minimal Unsatisfiable Subformula}).

\subsection{Recursive Model Rotation}

The \emph{model rotation} approach (\mr) has been introduced in
\cite{joao2011}. It is based on the \emph{transition assignment}
concept.

\begin{definition}[Transition assignment]
  Let $\PXC$ be a core.  An assignment $\A$ of $\P$ is a transition
  assignment of $\P$ iff $\A$ falsifies only one constraint in $\P$.
\end{definition}

\begin{property}
  Let $\PXC$ be a core and $c \in \P$. $c$ is a transition constraint
  of $\P$ iff there exists an transition assignment of $\P$ that
  falsifies $c$.
\end{property}

The proof is straightforward since $c$ is a transition constraint of
$\PXC$ iff $\P$ is unsatisfiable and there exists a solution $\A$ of
$\langle\X,\C\setminus\{c\}\rangle$ iff $\A$ falsifies only the
constraint $c$ of $\XC$. %\hfill$\Box$

When a transition assignment $\A$ is found, the model rotation
approach explores assignments that differ from $\A$ w.r.t.  only one
value. If this close assignment also falsifies only one constraint
$c'$, then $c'$ belongs to every \muc, too.

In \cite{Belov2011}, it is proposed to recursively perform model
rotation.  This extended technique is called \emph{Recursive Model
  Rotation}: it is summarized in a \csp version in Algorithm
\ref{alg:recursiveModeleRotation}.  This algorithm always makes local
changes to the value of one variable in the transition assignment in
trying to find out another transition assignment exhibiting another
constraint (lines 3--5), without a call to a constraint network
solver.  Contrary to the initial model rotation technique, the process
is not stopped when a transition assignment does not deliver an
additional transition constraint.  Instead, model rotation is
recursively performed with all transition assignments found (lines
6--8).  See e.g., \cite{Belov-Joao:12}  \cite{BLM:12}  and \cite{Wie11}  for more on the use of model rotation to
extract \mus{es}.

\restylealgo{ruled}\linesnumbered
\begin{algorithm}[t]
  \caption{\label{alg:recursiveModeleRotation} Recursive-\mr  (\mr stands for Model Rotation)}
  \SetLine
  \SetVline
  \SetInd{0.3em}{0.7em}
  \KwIn{$\PXC$: an unsatisfiable \cn, \\
    \hspace*{0.8cm} 
    $\C_{\muc}$: a set of constraints belonging to every %\textcolor{red}{every
    % ?} --> OK [Bertrand]
    \muc of $\P$, \\ \hspace*{0.8cm} $\A$: a transition assignment of $\P$}
  \KwOut{expanded $\C_{\muc}$}
  { 
    $c  \affect$ the  only constraint  falsified by  $\Ass$\tcc*{transition constraint}
    $\C_{\muc} \affect \C_{\muc}\cup\{c\}$\;
    \ForEach{$x \in \scp(c)$}
    {
      \ForEach{$v \in \dom(x)$}
      {
        $\Ass' \affect \Ass$ where the variable $x$ is assigned to $v$\;
        \If{$\Ass'$ is a transition assignment of $\P$}
        {
          Let  $c'$ be the  transition constraint  associated  to $\Ass'$
          w.r.t. $\P$\;
          \lIf{$c' \notin \C_{\muc}$}
          {
            $\C_{\muc} \affect$ Recursive-\mr{}($\P$, $\C_{\muc}$, $\Ass'$)\;
          }
        }
      }
    }

    \Return $\C_{\muc}$\;
  }
\end{algorithm}

\section{Local Search for Transition Constraints}

In the following, we introduce a new approach for computing one \muc by means of exhibiting transition
constraints  that relies on stochastic local search (in short, \SLS) as a
kernel procedure.  Whenever \SLS reaches an assignment that falsifies
exactly one constraint $c$, $c$ is a transition constraint and belongs to the
final \muc.  Obviously, such an assignment is a local minima for \SLS,
which can then explore neighborhood assignments, including other
possible transition assignments that would be discovered by recursive
model rotation.  Hence, we have investigated a generic approach based
on \SLS that we call \textit{Local Search for Transition Constraints}, in short \LSTC.

Local search in the \sat and \csp domains  is usually implemented as a tool intended to search for a model.
On the contrary, we make use of \SLS to explore the neighborhood of a
transition assignment in search for additional transition constraints.
Accordingly, some adaptations were made to the usual \SLS scheme.
Although its escape strategy is close to the so-called
\textit{breakout} method \cite{Morris1993}, the end criterion was
modified in order to allow SLS to focus and stress on parts of the
search space that are expectedly very informative, as proposed in
\cite{GLM-11-1,ALM+-10-1}.  More precisely, the $nbIt$ counter of
iterations remaining to be performed is increased in a significant way
each time an additional transition constraint has been discovered, as
SLS might have reached a promising part of the search space that has
not been explored so far. On the contrary, when an already discovered
transition constraint $c$ is found again, the $nbIt$ counter is
decreased by the number of times $c$ has already been considered, as a
way to guide SLS outside expectedly well-explored parts of the search
space. Otherwise, the $nbIt$ counter is decremented at each step and
the procedure ends when $nbIt$ becomes negative.

The objective function was itself modified to enforce the satisfaction
of the constraints already identified as belonging to the \muc that
will be exhibited.  More precisely, these latter constraints have
their weighs increased in order to be satisfied first.

Algorithm \ref{alg:SLSComponent} summarizes the approach.  It takes as
input an unsatisfiable constraint network $\P'=\langle
\X',\C'\rangle$, an assignment $\A$ and the current set of constraints
$\C_{\muc}$ that have already been recognized as belonging to the \muc
that will be exhibited.  Note that in most calls to Algorithm
\ref{alg:SLSComponent}, the $\P'$ parameter is a subpart of the \CN
$\P$ for which a \muc must be found; $\P'$ will represent the current
result of a dichotomy destructive strategy in the calling procedure.

\restylealgo{ruled}\linesnumbered
\begin{algorithm}[t]
  \caption{\label{alg:SLSComponent} \LSTC{ } (stands for Local Search for Transition Constraints)}
  \SetLine
  \SetVline
  \SetInd{0.3em}{0.7em}
  \KwIn{$\P'=\langle\X',C'\rangle$: a  \cn,  \\
    \hspace*{0.8cm} 
    $\C_{\muc}$: a set of constraints belonging to every %\textcolor{red}{every
    % ?} --> OK [Bertrand]
    \muc of $\P'$, \\ \hspace*{0.8cm} $\A$: a transition assignment of $\P'$
    (possibly empty)}
  \KwOut{expanded $\C_{\muc}$}
  { 
    \BlankLine
    \lIf{$\A=\emptyset$}{$\A \affect$ a random assignment of $\X'$\;}
    \lForEach{$c \in \C'$}
    {
      $w(c) = 1$\;
    }
    Initialize $nbIt$ by a preset positive number     \tcc*{Counter
      nbIt of remaining\newline \null\hfill iterations is initialized.}

    % \BlankLine
    \While{$(nbIt \geq 0)$}
    {            
      \SetJMline
      \If{a local minimum is reached}
      {             
        \SetVline   
        {\color{cyan}
          \If(\tcc*[f]{transition assignment}){$|\{c \in \C' | c$ is falsified by
            $\Ass\}| = 1$}
          {
            Let $c$ be the constraint falsified by $\Ass$ \tcc*{transition constraint} 
            \eIf{$c \notin \C_{\muc}$}
            {
              $\C_{\muc} \affect \C_{\muc} \cup \{c\}$\;
              Increase $nbIt$ by a preset positive bonus \;
            }
            {
              $nbIt \affect nbIt -  w(c)$ \;
              $w(c) \affect w(c) + 1$ \;                    
            }
          }}
        \SetVline
        Change the value in $\Ass$ of one var. of $\X'$ according to an
        escape strategy \;
      }      
      \lElse{
        Change the value in $\Ass$ of one var. of $\X'$ s.t. the sum of the weighs of violated constraints
        decreases\;
      }
      $nbIt \affect nbIt - 1$\;
    }
    \Return $\C_{\muc}$\;
  }
\end{algorithm}

Also, $\A$ does not need to be a transition assignment. When $\A$ is
empty, it is randomly initialized, like in a classical \SLS
procedure. Otherwise, $\A$ is a transition assignment which is used as
the starting point of the search.  The algorithm returns $\C_{\muc}$
after this set has been possibly extended by additional constraints
also belonging to this \muc.  The local search is a standard basic
random-walk procedure \cite{SelmanKC94} where the objective function
has thus been modified in order to take a specific weigh on each
constraint into account. Note that these weighs are specific to the
call to \LSTC{}  and thus different from the counters delivered by the
pre-processing step, which are used by the dichotomy strategy.  

A local minimum of a \SLS algorithm  is a state where there
  does not exist any assignment that can be reached by a single move
  of the local search and that would decrease the sum of the weighs of
  the falsified constraints.
Each time a local minimum is reached, the method tries to collect
information (lines 6--13). Thus, before applying an escape criterion
(line 14), when there is only one constraint $c$ of $\C'$ that is
falsified by the current assignment (i.e., when the current assignment
is a transition assignment), $c$ must appear in the final \muc. Two
sub-cases are thus as follows. When $c$ does not already belong to $
\C_{\muc}$, $c$ is inserted within $\C_{\muc}$ (line 9) and the value
of $nbIt$ is increased (line 10).  When $c$ already belongs to
$\C_{\muc}$, a penalty under the form of a negative number is applied
to $nbIt$ (line 12) and the weigh of $c$ is incremented (line 13).  In
this way, the more a transition constraint is considered, the greater
is the penalty. When SLS is not reaching a local minimum, the value of
one variable of $\X'$ is changed in such a way that the sum of the
weighs of the falsified constraints decreases (line 15).  In both
latter cases, the value of the $nbIt$ counter is decreased at each
loop (line 16).  Finally, when $nbIt$ reaches a strictly negative
value, a set of constraints included in the final \muc to be exhibited
is returned (line 17).

Let us stress that Algorithm \ref{alg:SLSComponent} without colorized
lines (6 to 13) is a mere standard stochastic local search procedure.

Noticeably, this approach differs from \cite{GMP06a,GMP-07-2} where a different form
of  \SLS was used to extract \mus{es}.  First,  \cite{GMP06a,GMP-07-2}  was dedicated to the Boolean
case using specific features of the clausal Boolean framework: extending it to the general constraint networks  setting while 
still obtaining 
acceptable running times for many real-life instances  remains an open challenge.  Second, it was used as a fast-preprocessing step
to locate an upper-approximation of a \mus. The role of \LSTC{} is  different: this procedure will be called during  the  fine-tuning process  of the approximation
delivered by the pre-processing.  Finally,
 \cite{GMP06a} and \cite{GMP-07-2} were based on the so-called \textit{critical clause} concept 
  to
 explore the search space, which is not generalized and adopted here in the general  framework of constraint networks.  

\section{Dichotomy Core Extraction with Local Search}

\restylealgo{ruled}\linesnumbered
\begin{algorithm}[t]
  \caption{\label{alg:coreExtraction} Dichotomy Core Extraction with
    Local Search ({\footnotesize \dcwcore+\LSTC})\hspace*{-2cm}\null}
  \SetLine
  \SetVline
  \SetInd{0.3em}{0.7em}
  \KwIn{$\PXC$: an unsatisfiable \cn}
  \KwOut{one \muc of $\P$}
  {
    \BlankLine
    $\C \affect \wcore(\CNLR{\X}{\C})$ \tcc*{Preprocessing step} 
    $\C_{\muc}\affect \emptyset$\tcc*{Set of constraints  belonging to the \muc} 
    {\color{cyan}$\ALS \affect \emptyset$ \tcc*{Last transition assignment found}}
    $\Ccut \affect$ choose $\lceil\frac{|\C|}{2}\rceil$ constraints of
    $\C$\footnotemark\tcc*{Set of constraints analyzed  \newline 
      \null\hfill for possible removal, selected according to  \newline\null\hfill  a dichotomy strategy}
    % \BlankLine
    \While{$\Ccut \ne \emptyset$}
    {
      $\Ass \affect solve(\CNLR{\X}{\C\setminus\Ccut})$
      \tcc*{Usual \mac algorithm is used}
      \If{$\Ass \neq \emptyset$ {\bf and} $|\Ccut| > 1$\hspace*{-0.1cm}
        \tcc*[h]{\hspace*{-0.12cm}A solution is found and more than one\hspace*{-0.12cm}\null}\newline} 
      {\vspace*{-0.4cm}\tcc*[f]{\hspace*{-0.17cm}constraint has been removed
          \hspace*{-0.17cm}\null}\newline
        $\Ccut \affect$ choose $\lceil\frac{|\Ccut|}{2}\rceil$
        constraints of
        $\Ccut$\addtocounter{footnote}{-1}\footnotemark\tcc*{\hspace*{-0.17cm}range of analyzed  \newline\null\hfill   constraints 
          is reduced\hspace*{-0.17cm}\null} 
      }
      \Else
      {
        \lIf{$\Ass = \emptyset$} 
        {
          $\C \affect \C \setminus \Ccut$ \tcc*[j]{No solution found,
            $\Ccut$ can be \newline\null\hfill
            removed from $\C$ while the resulting $\C$ remains unsat}
        }
        \Else{\vspace*{-0.4cm}\tcc*[f]{Solution found and $|\Ccut| = 1$}\newline
          $\C_{\muc} \affect \C_{\muc} \cup \Ccut$\tcc*{            A trans. const.
            has been found}
          {\color{cyan}$\ALS \affect \Ass$ \tcc*{The last
              transition assignment is saved}}
        }
        {\color{cyan}$\C_{\muc} \affect$ 
          \LSTC$(\CNLR{\X}{\C}, \C_{\muc}, \ALS)$\tcc*{$\C_{\muc}$ is
            extended by \LSTC}}
        {$\Ccut \affect$ choose
          $\lceil\frac{|\C\setminus\C_{\muc}|}{2}\rceil$ constraints of $\C\setminus\C_{\muc}$\addtocounter{footnote}{-1}\footnotemark}\;
      }
    }
    \Return $\langle \X,\C\rangle$\;
  }
\end{algorithm}

Algorithm \ref{alg:coreExtraction} summarizes an algorithm that
computes one \muc of an unsatisfiable constraint network $\PXC$, based
on the dichotomy strategy and relying on the local search procedure to extract additional transition constraints.
  As a preprocessing step, \wcore delivers in $\C$
an unsatisfiable core of $\P$ that is not guaranteed to be minimal
(line 1).  $\C_{\muc}$, the set of constraints that have already been
recognized as belonging to the \muc, is then initialized to the empty
set (line 2).  The lastly discovered transition assignment $\ALS$ is
initialized to the empty set (line 3).  According to a dichotomy
strategy $\Ccut$ is initialized with half the constraints of $\C$,
themselves selected according to the $dom/wdeg$ scores collected
during the \wcore preprocessing step (line 4). A dichotomy-based loop
is run until $\Ccut$ becomes empty.  While there remain constraints in
$\Ccut$ that have not yet been either removed from the candidate
\muc or inserted in this \muc, the sub-network $\CNLR{\X}{\C \setminus
  \Ccut}$ is solved and the solution is stored in $\Ass$ (line 6).  By
convention, when $\Ass$ is a model of $\CNLR{\X}{\C \setminus \Ccut}$,
it is not empty. In this case and when at the same time the number of
constraints belonging to $\Ccut$ is different from $1$, no conclusion
can be made with this set of constraints and $\Ccut$ is then refined
according to the dichotomy strategy.  When $\Ass$ is empty,
$\CNLR{\X}{\C \setminus \Ccut}$ is unsatisfiable and the constraints
of $\Ccut$ can be removed from $\C$ while keeping the unsatisfiability
of this latter set (line 9).  Finally, when $\Ass$ is not empty and
$\Ccut$ contains only one constraint $c$, $c$ is a transition
constraint and will appear in the final result $\C_{\muc}$ (line 10)
while the transition assignment is recorded in $\ALS$.

Calls to \LSTC{} are performed at each iteration step when $\Ass =
\emptyset$ or $|\Ccut| = 1$, with the following parameters: the
current \CN $\CNLR{\X}{\C}$, the set of constraints already identified
as belonging to $\C_{\muc}$ in construction and the complete
interpretation $\ALS$.  These calls are thus intended to find out
additional transition constraints with respect to $\C$.  Note that
after the first main iterations in the loop of Algorithm 3 have
allowed a first transition assignment to be found, all calls to \LSTC{} 
are made with the lastly discovered transition constraints as a
parameter.  Although there can thus exist several calls to \LSTC{} with
the same transition constraint, it is important to note that $\C$
evolves, forming another constraint network at each call.  It is thus
transmitted to the local search procedure together with an expectedly
``good" interpretation to start with.  When $\Ccut$ becomes an empty
set, this means that all constraints of $\C$ have been proved to
belong to the \muc, i.e., $\C=\C_{\muc}$.  Thus, at the end of the loop
$\C$ identified as forming one \muc of $\P$ is returned (line 14).

\footnotetext{The  $dom/wdeg$ scores collected during the \wcore step are
  used to rank-order  constraints.}

Importantly, Algorithm 3 is complete in the sense that it is
delivers one \muc for any unsatisfiable $\P=\CNLR{\X}{\C}$
in finite time (but it is exponential-time in worst case scenarios,
like all complete algorithms to find out one \muc).

Let us stress that this algorithm differs from the dichotomy
destructive strategy \dcwcore\mbox{ } in \cite{Hemery2006}
according to the colorized lines (namely lines 3, 11
and 12), only.  

%Algorithm 3 can be easily transformed to implement
%other forms of destructive approaches or \textsc{QuickXplain}-like
%refinement strategies \cite{junker04,Joao2013}, by adapting the way
%$\Ccut$ evolves.  For example, replacing
%$\lceil\frac{|\Ccut|}{2}\rceil$ by $1$ yields a destructive approach.

%%%%%%%%%%%%%%%%%%%%%%%%%%%%%%%%%%%%%%%%%%%%%%%%%%%%%% 

\section{Experimental Results}

In order to assess and compare the actual efficiency of the
local-search approach with other methods,  and in particular the model-rotation one, we have
considered all the benchmarks from the last \CSP solvers competitions
\cite{CSP08,CSP09},\footnote{The benchmarks are available at
  {\footnotesize
    \url{http://www.cril.univ-artois.fr/~lecoutre}}\hspace*{-3cm}\null}
which include binary vs.  non-binary, random vs.  real-life,
satisfiable vs.  unsatisfiable \CSP instances.  Among these instances,
only the 772 benchmarks that were proved unsatisfiable in less than
300 seconds using our own C++~ \rclcsp\footnote{\label{foot1} The
  executable is available at {\footnotesize
    \url{http://www.cril.univ-artois.fr/~lagniez}}} \CSP solver were
considered.  \rclcsp implements MAC embedding AC3 and makes use of the
$dom/wdeg$ variable ordering heuristic.

Three approaches to the \muc extraction problem have been implemented
with \rclcsp as kernel and experimentally compared: (1)
\dc{(\wcore{})} \cite{Hemery2006}, namely a dichotomy destructive
strategy with \wcore as a preprocessing step, without any form of
model rotation or local search to find out additional tran\-si\-tion cons\-traints,{} (2) \dc{(\wcore{})} + \rmr{} and~(3)
\dc{(\wcore{})}+\LSTC.
% , which  involve the recursive model rotation and he local-search model
% rotation, respectively.   
For the latter approach, the $nbIt$ counter was initialized to $10000$
and the bonus was set to that value, too. Furthermore, $rnovelty$
\cite{McAllesterSK97} was used as local-search escape criterion and
the advanced data structures proposed in \cite{LGM-12-1} have been
implemented.  All three versions have been run on a Quad-core Intel
XEON X5550 with 32GB of memory under Linux Centos 5.  Time-out has
been set to 900 seconds.

\begin{figure}[t]
  \centering
  \includegraphics[width=11cm]{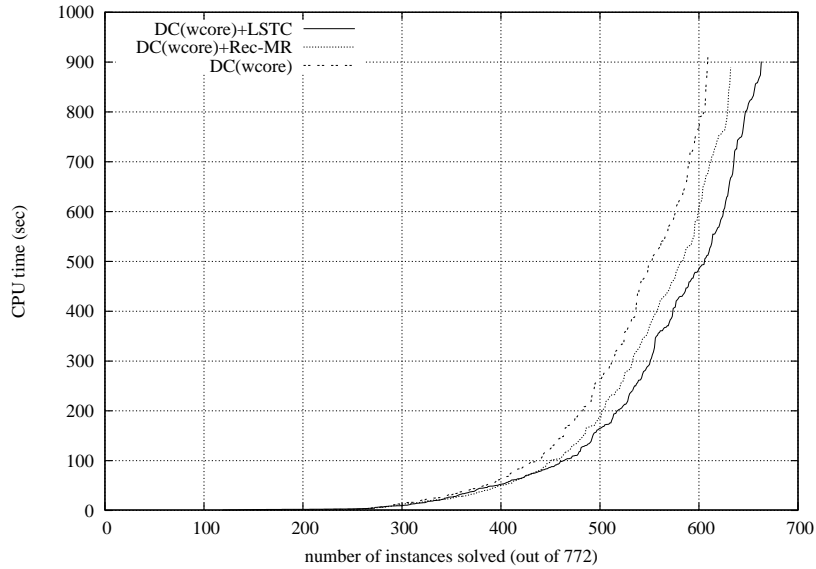}
  \caption{\label{fig:cactus}  Number of instances for which a \muc  was extracted.}
\end{figure}

In  Fig. \ref{fig:cactus}, a cactus plot compares the three approaches in
terms of the number of instances for which a \muc was extracted, indicating the 
spent CPU time on the $y$-axis.
Several observations can be made. First, similarly to the \sat
setting for model rotation \cite{Belov2011}, recursive model rotation improves 
performance  in the sense that \dc{(\wcore{})}+\rmr{} found a \muc for  632 
instances whereas
\dc{(\wcore{})} solved 609 instances, only. Then,  local search
 in its turn improves recursive model rotation according to the same criterion:
\dc{(\wcore{})}+\LSTC{}  solved 663 instances.  In addition to solving  54 and 31 additional 
instances respectively,  \dc{(\wcore{})}+\LSTC{} appears  to be more
efficient in terms of CPU time and less demanding  than the other competing approaches 
in terms of the number of calls to a \CSP solver for the more challenging  instances.

\begin{figure}[H]
  \centerline{\bf\hspace*{1.8cm}CPU times in sec.\hfill Number of calls
    to a \csp solver~~~~~~~~~~}\vspace*{-0.1cm}
  \centering
  \subfigure[\dc{(\wcore{})} vs. \dc{(\wcore{})}+\rmr{}]
  {%
    \label{fig:runningrmr}
    \hspace*{-0.45cm}\includegraphics[width=6.25cm]{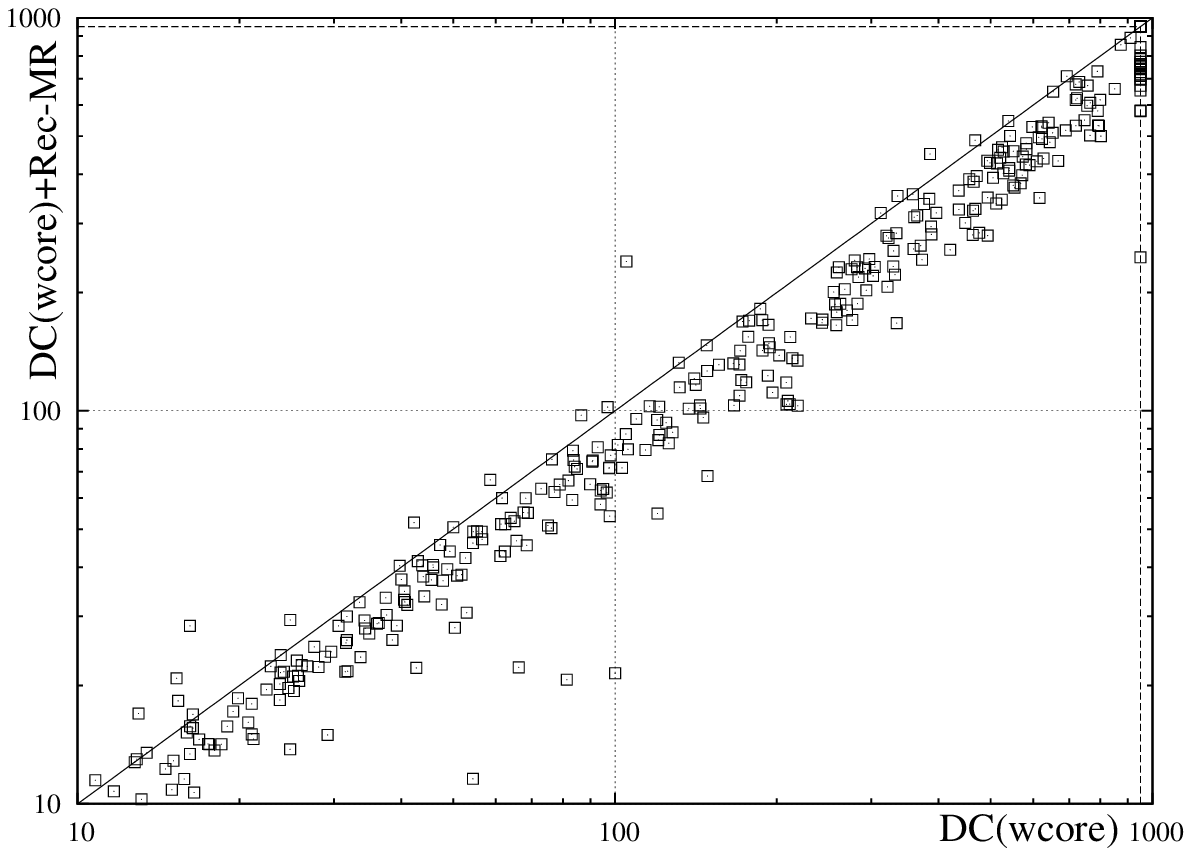}
  }% 
  \subfigure[\dc{(\wcore{})} vs. \dc{(\wcore{})}+\rmr{}]
  {%
    \label{fig:nbCallrmr}
    \includegraphics[width=6.25cm]{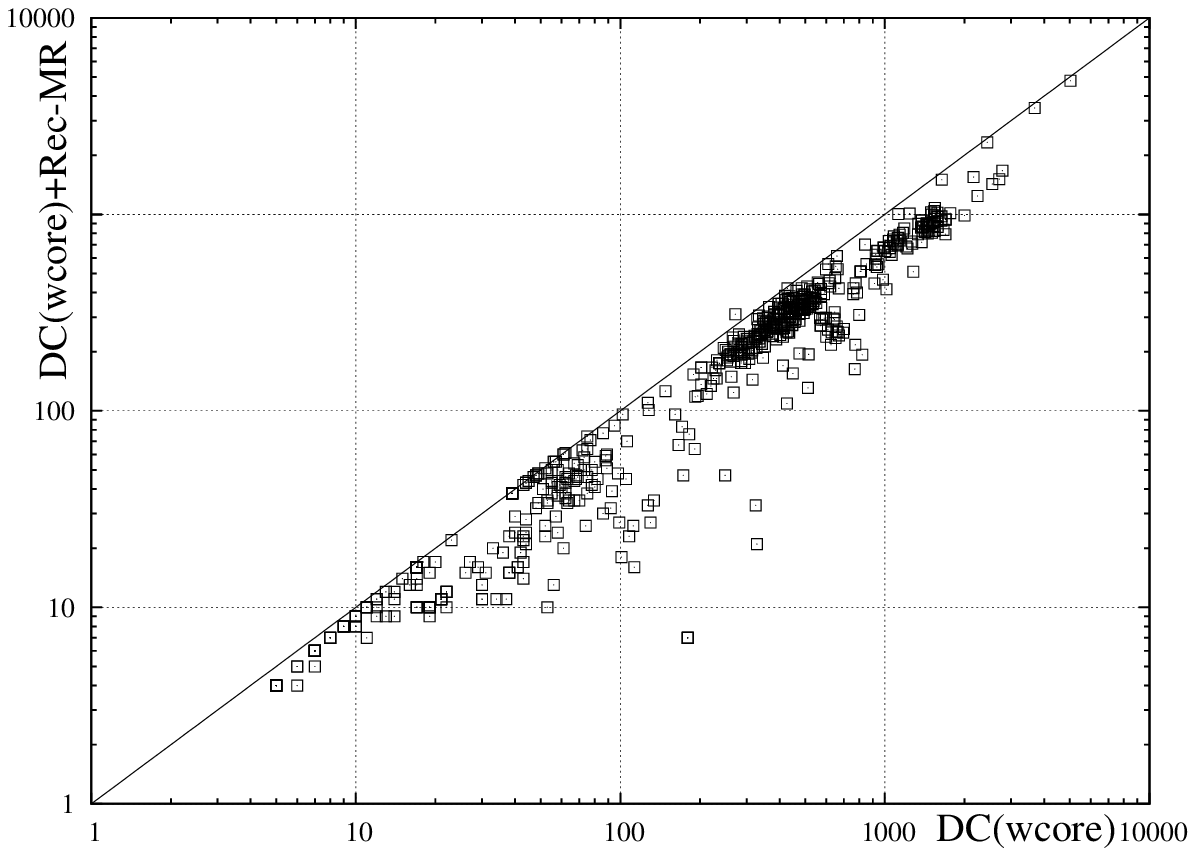}
  }% 
  \vspace*{-0.2cm}\\
  \subfigure[\dc{(\wcore{})} vs. \dc{(\wcore{})}+\LSTC{}]
  {%
    \label{fig:runninglsmr}
    \hspace*{-0.45cm}\includegraphics[width=6.25cm]{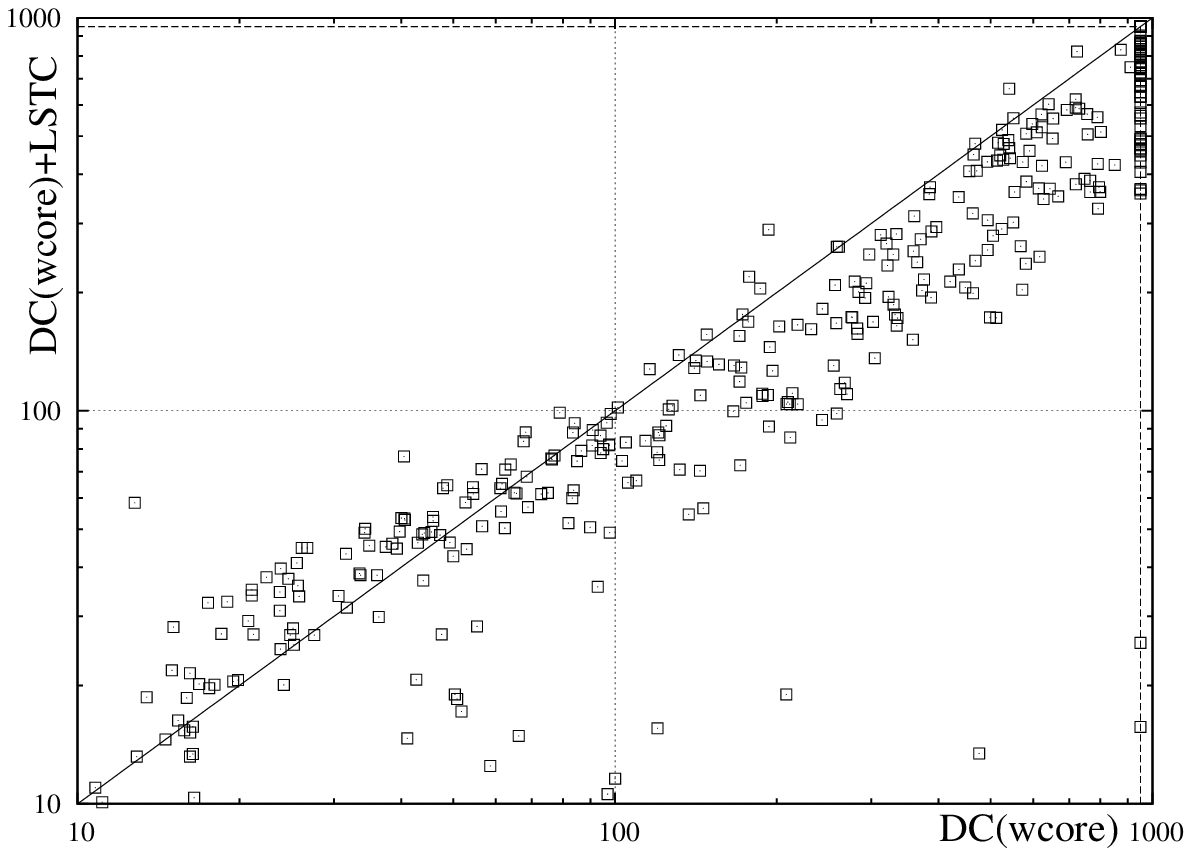}
  }%
  \subfigure[\dc{(\wcore{})} vs. \dc{(\wcore{})}+\LSTC{}]
  {%
    \label{fig:nbCalllsmr}
    \includegraphics[width=6.25cm]{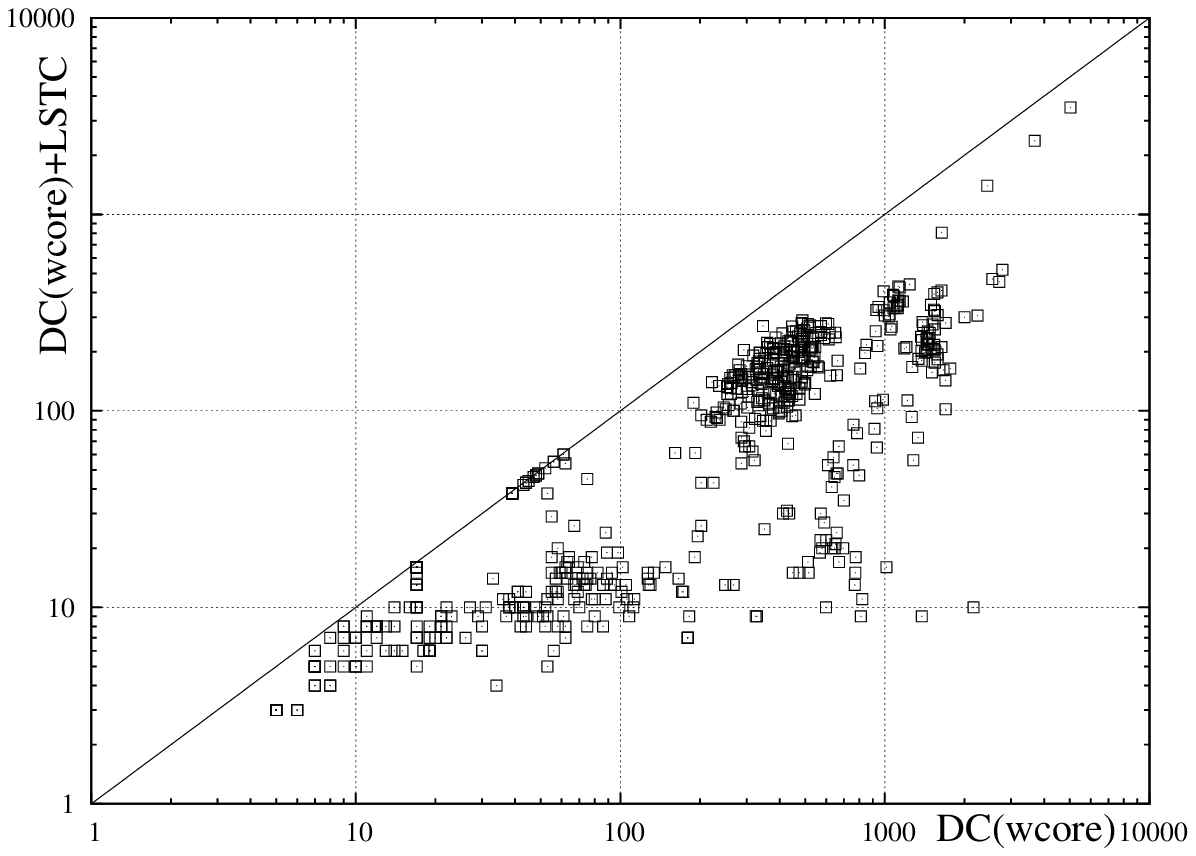}    
  }%
  \vspace*{-0.2cm} \\
  \subfigure[\hspace*{-0.1cm}\dc{(\wcore{})}+\rmr{} $\!$vs.$\!$ \dc{(\wcore{})}+\LSTC{}\hspace*{-0.5cm}\null]
  {
    \label{fig:runningrmrlsmr}
    \hspace*{-0.45cm}\includegraphics[width=6.25cm]{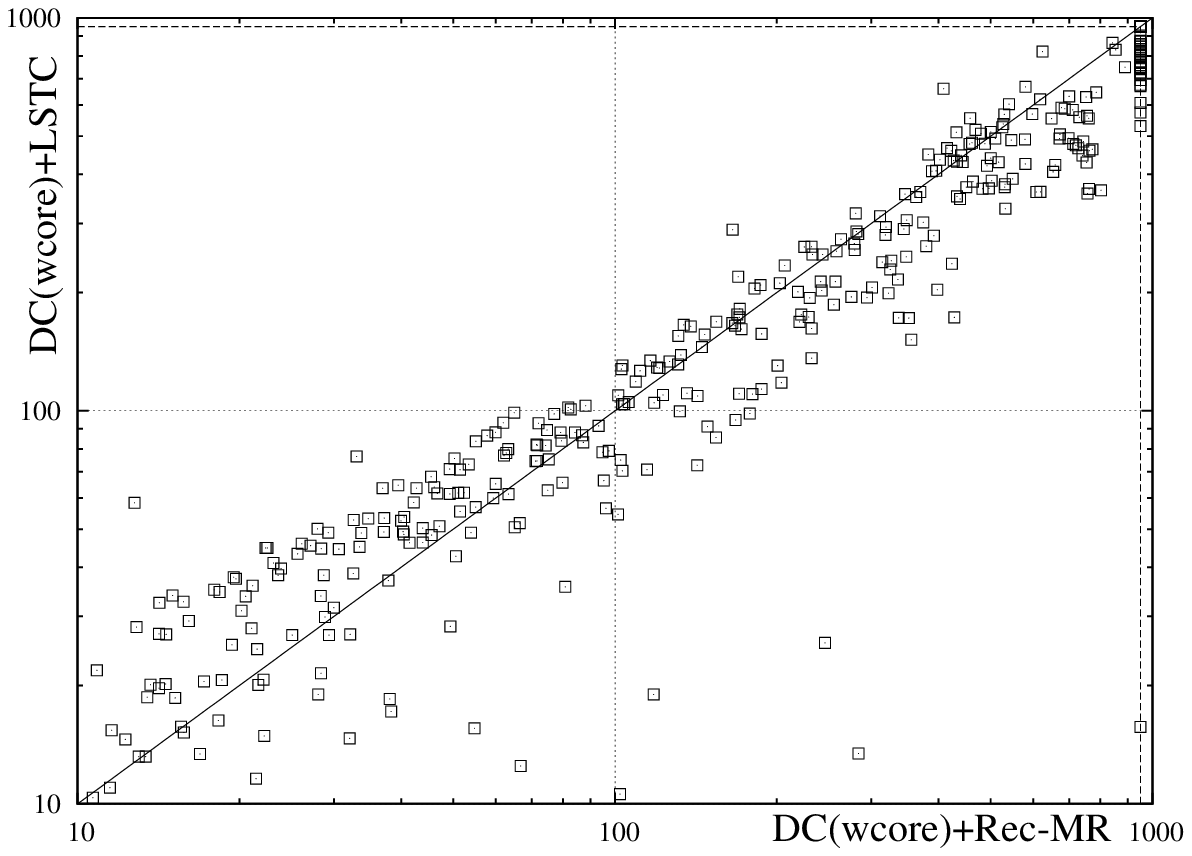}
  }%
  \subfigure[\hspace*{-0.1cm}\dc{(\wcore{})}+\rmr{} $\!$vs.$\!$ \dc{(\wcore{})}+\LSTC{}\hspace*{-0.5cm}\null]
  {
    \label{fig:nbCallrmrlsmr}
    \includegraphics[width=6.25cm]{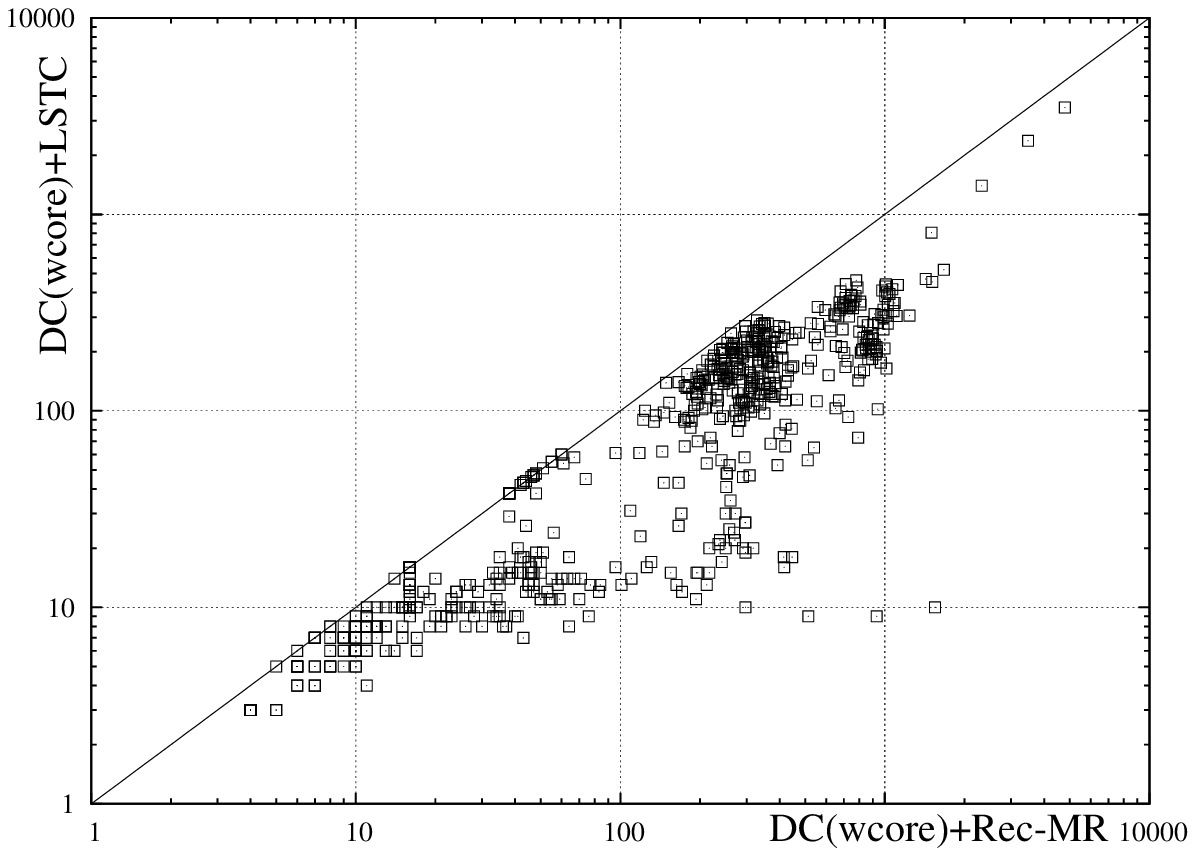}
  }%
  \vspace*{-0.3cm}\null
  \caption{      \label{fig:scatterTime}  Pairwise comparisons
    of \dcwcore{}, \dcwcore+\rmr and \dcwcore+\LSTC.} 
  \vspace*{-0.7cm}\null
\end{figure}

In Fig.~\ref{fig:scatterTime}, pairwise comparisons between the
approaches are provided.  Each comparison between two methods is done
in terms of CPU time (sub-figures (a) (c) and (e)) and of the number
of calls to the MAC solver (sub-figures (b), (d) and (f)),
successively.  For each scatter, the $x$-axis (resp. $y$-axis)
corresponds to the CPU time $tx$ (resp. $ty$) obtained by the method
labelled on the same axis. Each dot ($tx$, $ty$) thus gives the
results for a given benchmark instance. Thus, dots above (resp. below)
the diagonal represent instances for which the method labelled on the
$x$-axis is better (resp. worse) than the method labelled on the
$y$-axis. Points on the vertical (resp. horizontal) dashed line mean
that the method labelled on the $x$-axis (resp. $y$-axis) did not
solved the corresponding instance before timeout.

\begin{figure}[t]
  \centering 
  \begin{psfrags}
    \psfrag{lsmr}{{}}
    \psfrag{rmr}{{}}
    \hspace*{-0.45cm}\includegraphics[width=12cm]{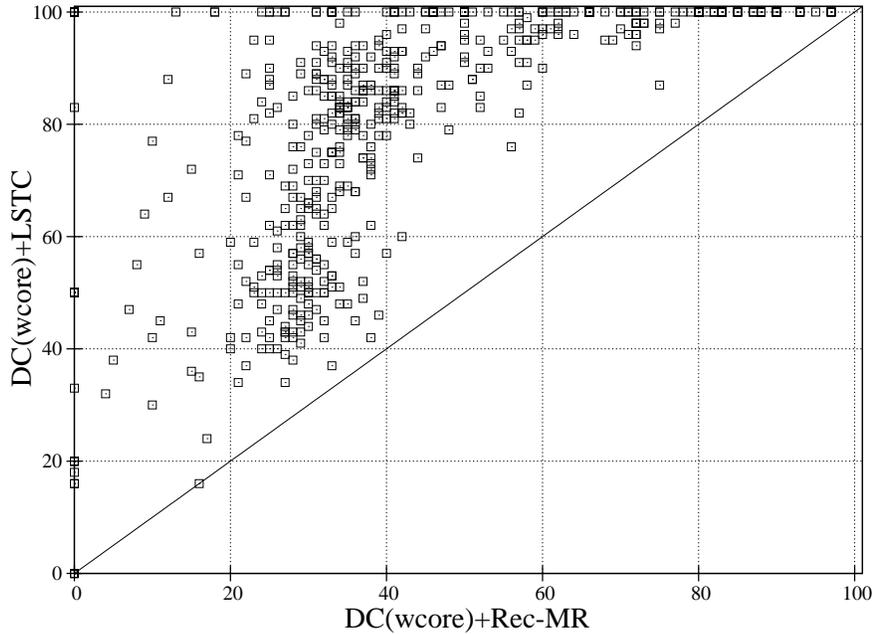}
  \end{psfrags}
\vspace*{-0.8cm}
  \caption{
    \label{fig:ratioTC}  $\%$ of the constraints in the \muc discovered by \rmr and \LSTC.}
\end{figure}

First, the figure \ref{fig:runningrmr} shows that \dc{(\wcore{})}+\rmr
solves instances faster than \dc{(\wcore{})}.  Then, Fig. 4(c) and
4(e) show that, most generally, \dc{(\wcore{})}+\LSTC{} finds one \muc
faster than both \dc{(\wcore{})} and \dc{(\wcore{})}+\rmr manage to do
it, provided that the instance is difficult in the sense that
extracting a \muc requires more than 100 seconds.  On easier instances,
the additional computational cost of local search (and model rotation)
has often a negative impact on the global computing time, but this
latter one remains however very competitive.  Fig. 4(b) (d) and (f)
show the extent to which recursive model rotation and local search
 reduce the number of calls to a \CSP solver, and thus
of calls to an NP-complete oracle, in order to find out one
\muc. Clearly, \dc{(\wcore{})}+\LSTC{} often outperforms the other
approaches in that respect.  The observation of these three figures
suggests that local search  allows more transition
constraints to be discovered by considering assignments in the
neighborhood of the transition assignments.  This intuition is
confirmed by the experimental results reported in
Fig.~\ref{fig:ratioTC}.  In this latter figure, we give the percentage
of the total size of the \muc that has been found by \rmr and \LSTC,
respectively.  It shows that \LSTC{} detects more transition constraints
than \rmr.  Moreover, it shows that for almost all instances, more
than half of the constraints in the \muc are found thanks to local
search when \dc{(\wcore{})}+\LSTC{}  is under consideration. This
ability explains much of the performance gains obtained on difficult
instances. Actually, local search detects the totality of the \muc for
many instances.

Finally, Tab.~\ref{tab:selectedInstances} reports the detailed 
results of each approach on a typical panel of instances from the benchmarks. 
The first 
four  columns provide the name, 
numbers of  constraints and variables of the instance, the number of remaining constraints after the preprocessing step, successively. Then, for 
each method, the CPU time, the size of the extracted  \muc  (the \muc{s} discovered by the various methods can differ) and 
the  number of \csp   calls to find  it are listed.  In addition, for 
\dc{(\wcore{})}+\rmr  (resp. \dc{(\wcore{})}+\LSTC),   the number (``by rot.")  
of constraints of the \muc detected by
model rotation (resp. local search  (``by LS")) is provided.
TO means time-out and 
the best computing time for each instance is shaded in grey.
For example,  \muc{s} of a same size (94 constraints) were found for the cc-10-10-2  instance using each of the three methods. 
\LSTC was best performing in extracting a \muc in 28.28 seconds (vs. 49.38 and 55.36 seconds for the other methods).
Note that all constraints in that \muc were discovered through the \LSTC{} procedure.

\begin{table}[H]
\vspace*{-1cm}
  \centering
  \renewcommand{\tabcolsep}{0.5pt}
\scalebox{.96}{
  \begin{tabular}{|c|c|c||c||c|c|c||c|c|c|c||c|c|c|c|}
    \hline
    \multicolumn{3}{|c||}{Instances} & Prep. &
    \multicolumn{3}{c||}{\dc{(\wcore{})}} &
    \multicolumn{4}{c||}{\dc{(\wcore{})}+\rmr} &
    \multicolumn{4}{c|}{\dc{(\wcore{})}+\LSTC} \\
    Name & $|\C|$ & $|\X|$ & $|\C'|$ & time & size & solver & time & size & by & solver & time & size & by & solver \\
    & & & & (sec) & \muc & calls & (sec) & \muc & rot. & calls & (sec) &
    \muc & LS & calls \\
    \hline
    radar-8-24-3-2 & 64 & 144 & 41 & 213.84 & 8 & 31 & 136.15 & 8 & 6 & 15 & \viewResul{110.90} & 8 & 7 & {10} \\
    radar-9-28-4-2 & 81 & 168 & 47 & 62.39 & 2 & 12 & \viewResul{43.86} & 2 & 1 & 10 & 50.28 & 2 & 2 & {8} \\
    \hline
    qKnights-50-add & 1235 & 55 & 1233 & 358.21 & 5 & 148 & 355.81 & 5 & 4 & 126 & \viewResul{151.64} & 5 & 5 & {16} \\
    qKnights-50-mul & 1485 & 55 & 1484 & 335.55 & 5 & 102 & 352.27 & 5 & 4 & 96 & \viewResul{172.16} & 5 & 5 & {16} \\
    qKnights-25-mul & 435 & 30 & 435 & 37.58 & 5 & 268 & 30.26 & 5 & 4 & 212 & \viewResul{9.19} & 5 & 5 & {13} \\
     qKnights-25-add & 310 & 30 & 309 & 22.88 & 5 & 127 & 22.38 & 5 & 4 & 110 & \viewResul{9.49} & 5 & 5 & {14} \\
     qKnights-20-add & 200 & 25 & 198 & 6.22 & 5 & 73 & 5.96 & 5 & 4 & 58 & \viewResul{3.69} & 5 & 5 & {13} \\
    \hline
    bdd-30 & 2713 & 21 & 1157 & TO & - & - & TO & - & - & - & \viewResul{855.96} & 10 & 3 & {59} \\
    bdd-2 & 2713 & 21 & 1307 & TO & - & - & TO & - & - & - & \viewResul{725.23} & 10 & 4 & {49} \\
    bdd-25 & 2713 & 21 & 1392 & 719.65 & 10 & 53 & 677.24 & 10 & 1 & 48 & \viewResul{590.79} & 10 & 3 & {38} \\
    bdd-18 & 2713 & 21 & 994 & TO & - & - & TO & - & - & - & \viewResul{824.97} & 10 & 6 & {46} \\
    bdd-6 & 2713 & 21 & 802 & TO & - & - & TO & - & - & - & \viewResul{530.59} & 9 & 4 & {28} \\
    bdd-3 & 2713 & 21 & 1551 & 910.64 & 11 & 75 & 889.34 & 11 & 0 & 74 & \viewResul{748.77} & 11 & 2 & {45} \\
    \hline
    ssa-0432-003 & 738 & 435 & 594 & TO & - & - & TO & - & - & - & \viewResul{801.17} & 307 & 301 & {265} \\
    \hline
    graph2-f25.xml & 2245 & 400 & 1364 & 19.48 & 43 & 841 & \viewResul{17.18} & 43 & 10 & 702 & 20.48 & 43 & 41 & {197} \\
    \hline
    qcp-10-67-10 & 822 & 100 & 511 & TO & - & - & TO & - & - & - & \viewResul{15.69} & 93 & 93 & {13} \\
    qcp-10-67-11 & 822 & 100 & 413 & 11.29 & 49 & 776 & 7.86 & 49 & 43 & 446 & \viewResul{1.99} & 49 & 49 & {18} \\
    \hline
    ruler-34-9-a4 & 582 & 9 & 276 & TO & - & - & 802.14 & 36 & 26 & 171 & \viewResul{364.23} & 36 & 36 & {12} \\
    ruler-17-7-a4 & 196 & 7 & 97 & 2.05 & 29 & 196 & \viewResul{1.78} & 29 & 16 & 119 & 8.49 & 29 & 29 & {23} \\
    ruler-25-8-a4 & 350 & 8 & 152 & 24.19 & 28 & 182 & 21.69 & 28 & 21 & 76 & \viewResul{20.08} & 28 & 28 & {9} \\
    \hline
    
    cc-10-10-2 & 2025 & 100 & 381 & 55.36 & 94 & 579 & 49.38 & 94 & 45 & 291 & \viewResul{28.28} & 94 & 94 & {20} \\
    cc-15-15-2 & 11025 & 225 & 902 & 275.42 & 92 & 572 & 229.33 & 92 & 47 & 272 & \viewResul{173.37} & 92 & 92 & {30} \\ 
    cc-12-12-2 & 4356 & 144 & 333 & 137.13 & 93 & 571 & 101.32 & 93 & 47 & 271 & \viewResul{54.49} & 93 & 93 & {22} \\
    cc-20-20-2 & 36100 & 400 & 1232 & TO & - & - & 755.35 & 92 & 43 & 297 & \viewResul{563.57} & 92 & 92 & {27} \\
    \hline
    ehi-85-297-14 & 4111 & 297 & 265 & \viewResul{1.43} & 38 & 428 & 1.54 & 37 & 11 & 420 & 2.19 & 37 & 33 & {113} \\
    ehi-85-297-64 & 4113 & 297 & 182 & 1.39 & 38 & 427 & \viewResul{1.06} & 37 & 13 & 319 & 1.86 & 35 & 28 & {105} \\
    ehi-85-297-98 & 4124 & 297 & 682 & 0.69 & 22 & 192 & \viewResul{0.47} & 22 & 11 & 118 & 1.09 & 21 & 20 & {61} \\
    \hline
    
    s-os-taillard-4-4 & 160 & 32 & 159 & 396.21 & 42 & 346 & 319.14 & 42 & 22 & 186 & \viewResul{293.55} & 42 & 35 & {89} \\
    s-os-taillard-4-9 & 160 & 32 & 160 & 322.84 & 37 & 302 & 275.31 & 37 & 13 & 232 & \viewResul{195.03} & 37 & 29 &{104} \\
    s-os-taillard-4-10 & 160 & 32 & 148 & 37.46 & 29 & 268 & \viewResul{33.46} & 29 & 12 & 190 & 45.05 & 29 & 25 & {100} \\
    s-os-taillard-5-25 & 325 & 50 & 279 & 186.26 & 10 & 38 & \viewResul{181.76} & 10 & 9 & 15 & 204.69 & 10 & 10 & {10} \\
    \hline
    BlackHole-1 & 431 & 64 & 142 & 105.03 & 75 & 658 & 240.09 & 80 & 56 & 269 & \viewResul{8.08} & 75 & 73 & {24} \\
    BlackHole-5 & 431 & 64 & 140 & 208.35 & 77 & 657 & 118.04 & 77 & 58 & 253 & \viewResul{18.98} & 82 & 80 & {48} \\
    BlackHole-0 & 431 & 64 & 139 & 96.92 & 75 & 645 & 102.16 & 77 & 56 & 291 & \viewResul{10.59} & 75 & 72 & {46} \\
    BlackHole-4 & 431 & 64 & 140 & 100.07 & 77 & 664 & 21.48 & 77 & 55 & 252 & \viewResul{11.59} & 77 & 74 & {48} \\
    BlackHole-3 & 431 & 64 & 142 & 476.00 & 80 & 700 & 283.92 & 80 & 58 & 261 & \viewResul{13.43} & 75 & 74 & {35} \\
    \hline
    
  \end{tabular}
}
  \vspace*{-0.3cm}\begin{flushleft}~The full table can be downloaded at {\scriptsize\url{http://www.cril.fr/~mazure/bin/fulltab-LS-MUC-GLM.pdf}}\hspace*{-25cm}~\end{flushleft}
  \vspace*{-0.2cm}\caption{\label{tab:selectedInstances} Some typical experimental
    results.}
\end{table}

\section{Perspectives and conclusion}

Clearly, the local search scheme proposed in this paper improves the extraction of one \muc by means of  destructive strategies and opens many perspectives.
Although dichotomy strategies, as explored in this paper,  are known to be the most efficient ones, it could be interesting to graft this  local search  
scheme to constructive or \emph{QuickXplain}-like 
methods.
Also, note  that we have not  tried to fine-tune the various parameters of this local search scheme.  In this respect, it would be interesting to devise forms of dynamical settings
for these parameters that better take the recorded information about  the previous search steps  into account, as explored in 
 \cite{GMP-08-4}.
 In the future, we plan to explore 
more advanced concepts that are related to  transition constraints in the goal of better guiding the local search towards  promising parts of the search space.
Especially, so-called critical clauses \cite{GMP-06-3} in the Boolean framework could be generalized in various ways in the full constraint networks  setting. Exploring the
possible  ways according to which 
 \LSTC{}  could benefit from this  is a promising path for further research.

\section*{Acknowledgements}

This work has been partly supported by a grant from the \textit{Région Nord/Pas-de-Calais} and by  an EC FEDER grant.

% -------------------------------------------------------------------------
\bibliographystyle{plain}
\bibliography{mabiblio}

% ------------------------------------------------------------------------- 
% \appendix

\end{document}